\documentclass{article}

\usepackage{arxiv}

\usepackage[utf8]{inputenc} 
\usepackage[T1]{fontenc}    
\usepackage{hyperref}       
\usepackage{url}            
\usepackage{booktabs}       
\usepackage{amsfonts}       
\usepackage{nicefrac}       
\usepackage{microtype}      
\usepackage{lipsum}
\usepackage{graphicx}
\graphicspath{ {./images/} }

\title{Path Planning and Obstacle Avoidance Scheme for Autonomous Robots using Raspberry Pi}

\author{
 R. N. Somarathna \\
  Department of Information Technology\\
  Faculty of Information Technology\\
  University of Moratuwa, Sri Lanka \\
  \texttt{nilakshikasomarathna@gmail.com} \\
}

\begin{document}
\maketitle
\begin{abstract}
With the incremental development of robotic platforms to automate the manual processes, path planning has become a critical domain with or without the knowledge of the indoor and outdoor environment. The algorithms can be intelligent or pre-structured and should optimally reach the destination efficiently. The major challenge in this domain is to find a path which is free from static obstacles as well as dynamic obstacles. In this paper, a methodology is proposed with the implementation details of the robotic platform to cover the critical key points and to arrive at the original key point in a dynamic environment. The main computation is happening inside a Raspberry Pi B+ module, and compass, wheel encoders and ultrasonic sensors were used in the implementation for the localization of the robot to relevant key points. 
\end{abstract}


\section{Introduction}
Path planning and obstacle avoidance in a dynamic environment are the problems of interest in the platform of robotic module generation. Path planning \cite{RN429} is the discovery of routes that goes through the locations in the environment and is a set of location points in the area of interest that the robot needs to traverse. Path planning enables the determination of static obstacles in the environment and proposes a solution which is feasible in the static environment. Obstacle avoidance is the approach of detecting the obstacles in the environment and ability to evade them. Obstacle avoidance can be categorized into two types as static obstacle avoidance and dynamic obstacle avoidance. Static obstacle avoidance can be exercised by the path planning strategy and dynamic obstacle avoidance can be implemented by using the sensors to the robotic platform. Path planning with obstacle avoidance provides the ability to generate an optimal path to cover the environment with less cost. 

There are many application areas that the robot needs to cover the specific set of key points in a defined environment and arrive at the original location optimally and efficiently. More specifically this approach can be best used in the surveillance of shopping malls, guiding of the customers to shelves in shopping malls, finding a lost person in a crowded environment, and food delivery in restaurants. Consequently, the proposed solution will cater to all the above-mentioned scenarios and will optimally provide the full coverage. Given that, this research aims to provide a low resource consumption autonomous navigation module which can work in a dynamic environment.

The solution is configured via a Raspberry Pi which operates hardware platform with navigation ability using the ultrasonic sensors, compass, and encoders. The navigation module will roam through the key points and come to the original location for the next operation. The system can avoid a static and dynamic obstacle in the environment. 

Next section of the research will elaborate on the information about the design of the hardware platform, and the methodology section will have all the information related to the algorithm and implementation. Following that, result and discussion will illustrate the extent to which the objectives are achieved.

\section{Design}
\subsection{Python for the computation}
Python is extensively used general-purpose, high-level programming language. The language provides syntaxes which enable expressing concepts in fewer lines of code. Python will be used as the main programming language to build the computation engine which will handle the motion planning, navigation, and obstacle avoidance because of the presence of numerous libraries and ease of use.

\subsection{Raspberry Pi}
\begin{figure}[ht]
  \centering
  \includegraphics[width=10.5cm, height=6.5cm]{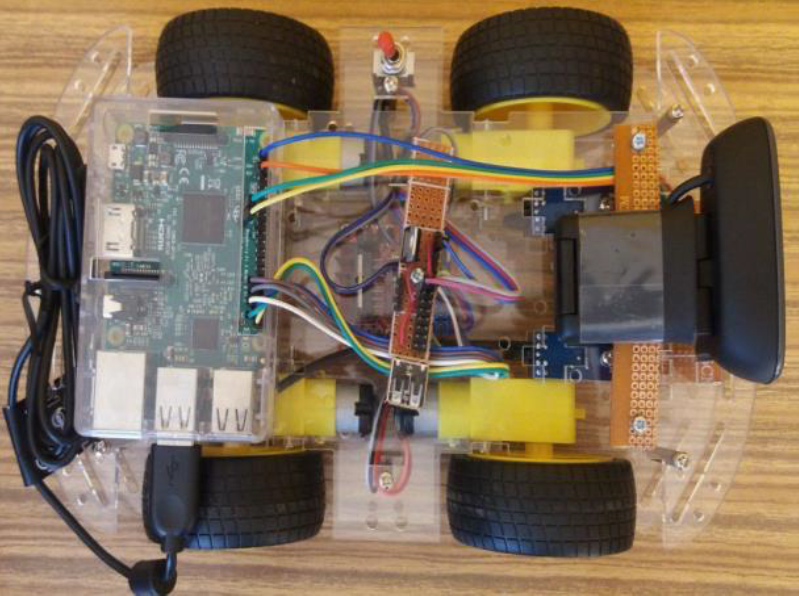}
  \caption{Hardware design of navigation module}
  \label{fig:Hardware}
\end{figure}

As shown in (Figure \ref{fig:Hardware}), we manipulated the research via a Raspberry Pi model. It is a single board, a high-performance computer which currently has different modules as Model A, Model B, Model A+, Model B+ \cite{RN430}. In our system, we used low power consumed Model B+ which has 512 MB ram with 4 Universal Serial Bus (USB) ports, USB/Ethernet chip with a five-point USB hub, and four ports.

\subsection{Ultrasonic sensors}
The ultrasonic sensor is a component that calculates the distance to an object with the use of sound waves. This interprets the reflected signals from objects and measures the time between sound wave generation and sound wave reception. These ultrasonic sensors are using to tackle the dynamic obstacles in the environment. In here the sensor will emit an ultrasonic wave when operating the robot. If there is a collision with an object it will reflect the sensor. If there is a reception to the sensor, the robot will assume the presence of an obstacle in front of it. Then the time difference between generation and reception of ultrasonic waves will be calculated \cite{RN430}. 

Distance to object = Speed x Time / 2 

\subsection{VNC viewer}
Virtual Network Computing (VNC) Viewer was used as the software to remotely control the Raspberry Pi over the network. A view of the Raspberry Pi is displayed in the local computer so can control the Raspberry Pi. This is platform-independent and works on any operating system.

\subsection{I2C libraries}
I2C or Inter-Integrated-Circuit bus is a serial interface in Raspberry Pi which support many devices until the addresses of each device do not conflict with each other. This bus provides the opportunity to share data among microcontrollers and components with a minimum of wiring. 

\subsection{Compass}
Compass model HMC5883L was used in the navigation to get the bearings of the directions and was connected to the Raspberry Pi. 

\subsection{Wheel encoders}
Wheel encoders are used to determine the forward distance that the module travel in a particular direction. Two-wheel encoders were connected to the front wheels in the implementation. Several ticks need for a particular displacement can be calculated by follows \cite{RN431}. 

Number of Revolutions = Distance/ Circumference

Tick count = Number of Revolutions x Counts per revolution 

\section{Methodology}
\subsection{Reception of area map, key points, and bearings}
\begin{figure}[ht]
  \centering
  \includegraphics[width=8.5cm, height=6.5cm]{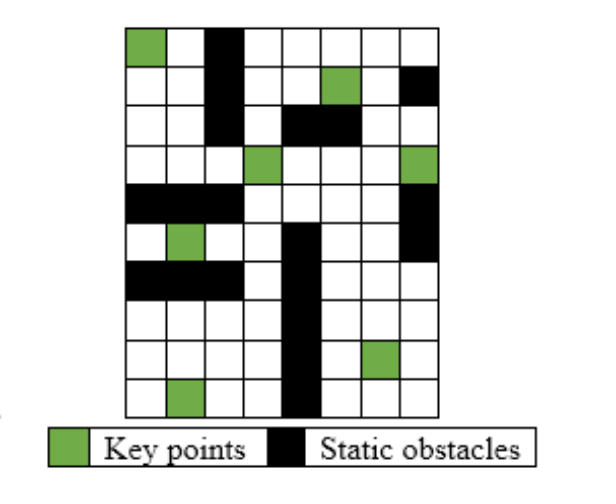}
  \caption{Map with key points}
  \label{fig:Map}
\end{figure}

\begin{figure}[ht]
  \centering
  \includegraphics[width=8.5cm, height=5.5cm]{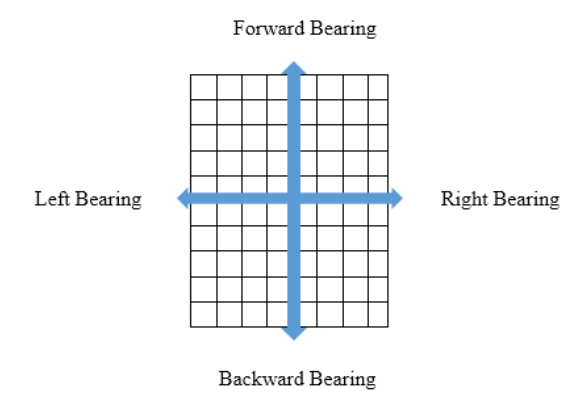}
  \caption{Bearings for the four main directions}
  \label{fig:Bearings}
\end{figure}

For the initialization of the localization of the robotic platform first need to retrieve the area map, assigned key points and bearings by each of the individual module. In this approach, the real world defined environment is mapped to a two-dimensional array (This research does not cover the mapping of the environment to the array). The two-dimensional array of the map will include the static obstacles as ‘1’ and the free spaces in the environment as ‘0’ (Figure \ref{fig:Map}). Then set of key points will be generated and given as inputs to each module. Key points are the critical areas which the module need to visit at least once and given as location values in an array. Bearings for the four directions as forwarding, backward, left, right need to be measured by the compass and should give as inputs to the module, because the module will support only the above four directions. The processing of getting bearings of the area is shown by the following diagram (Figure \ref{fig:Bearings}).

\subsection{Key points coverage strategy}
After getting the map of the area, the goal of the navigation module is to cover the set of key points optimally and reaches the original key point. Travelling Salesman Problem (TSP) algorithm \cite{RN432} can be used to cover all the key points optimally and reaches the original point. The problem of the domain is to cover all the key points using the shortest route with the least cost. 

Distance between one key point to another key point can be calculated by using the A-Star algorithm considering the static obstacles in the environment. Hence the path generated is always accessible in the static environment. In this algorithm, the estimated total cost of the path through “i” to goal F(i) was calculated by following equation, where g(i) is the path cost from the initial point to ‘i’ and h(i) is the estimated cost from “i” to the goal \cite{RN433}. 

F(i)= g(i)+h(i) 

\begin{figure}[ht]
  \centering
  \includegraphics[width=6.5cm, height=4.5cm]{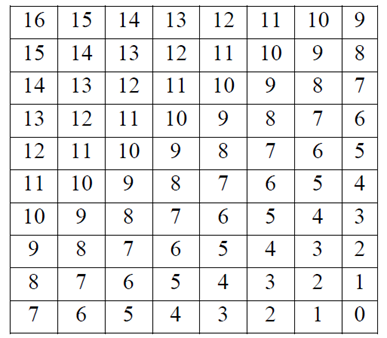}
  \caption{Heuristic allocation}
  \label{fig:Heuristic}
\end{figure}

In the algorithm, the heuristic assignment can be done by considering the goal as the (n,m) position of an [n x m] array \cite{RN433}. The procedure of allocating cost functions to each cell in the array does not consider the obstacles in the map. The cost function is calculated as the distance from the current location to the defined goal in the map by only considering the above-defined four directions. So, each cell has a value depicting the cost to move from its location to the goal taking the cost to move from one cell to another as one. The following figure shows the assignment of the heuristic to a 10x8 array taking the [10,8] cell as the goal (Figure \ref{fig:Heuristic}). 

\begin{figure}[ht]
  \centering
  \includegraphics[width=8.5cm, height=5.5cm]{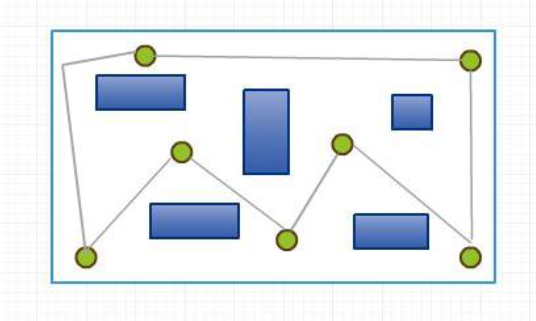}
  \caption{Key points coverage mechanism by TSP}
  \label{fig:coverage}
\end{figure}

In here only four directional movements were considered as forward, left, backward, and right through a 2D array, which is represented respectively as [[-1,0], [0, -1], [1,0], [0,1]]. The algorithm is starting at the given initial key point and it will traverse through the nodes until the goal key point is found (Figure \ref{fig:coverage}). If there are several free nodes available, after opening all those nodes the node with the smallest cost will be selected as the next move. Then the heuristic for the current location is calculated and this procedure will run until the goal key point is found. After reaching the goal total cost between key points is calculated. This process can be done using the simulated annealing to find the global minimum solution between each of the key points with others. So, can get the global minimum distance between each of the key points and the order of coverage of key points.

\subsection{Navigation through key points}
\begin{figure}[ht]
  \centering
  \includegraphics[width=7.5cm, height=5.5cm]{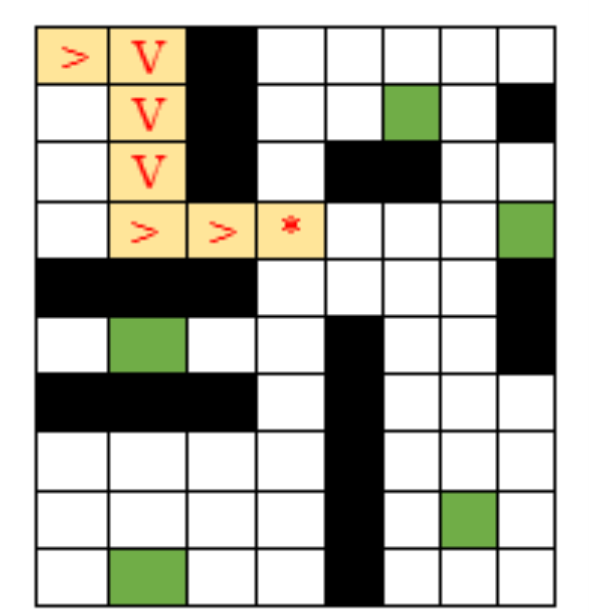}
  \caption{Path between two key points}
  \label{fig:PathPlan}
\end{figure}

After getting the order of the key points, the directions of the robotic path between one point to another need to be calculated (Figure \ref{fig:PathPlan}). This path can be calculated using the A-Star algorithm \cite{RN433}. In the algorithm, the array can be generated by using the values allocated by the above forward, backward, left, and right moves in the map. Then these move values can be mapped to the four bearing values in the real-world environment.

\subsection{Compass Guidance}
Compass model HMC5883L was used in the implementation of the navigation model to get the bearings of the directions. In the implementation of the module, bearings for the forward, left, right, and backward directions of the real-world area are measured and that was given as an input to each of the modules. So, the compass will be used as the guidance to turn the module into the relevant direction. 

\begin{figure}[ht]
  \centering
  \includegraphics[width=8.5cm, height=5.5cm]{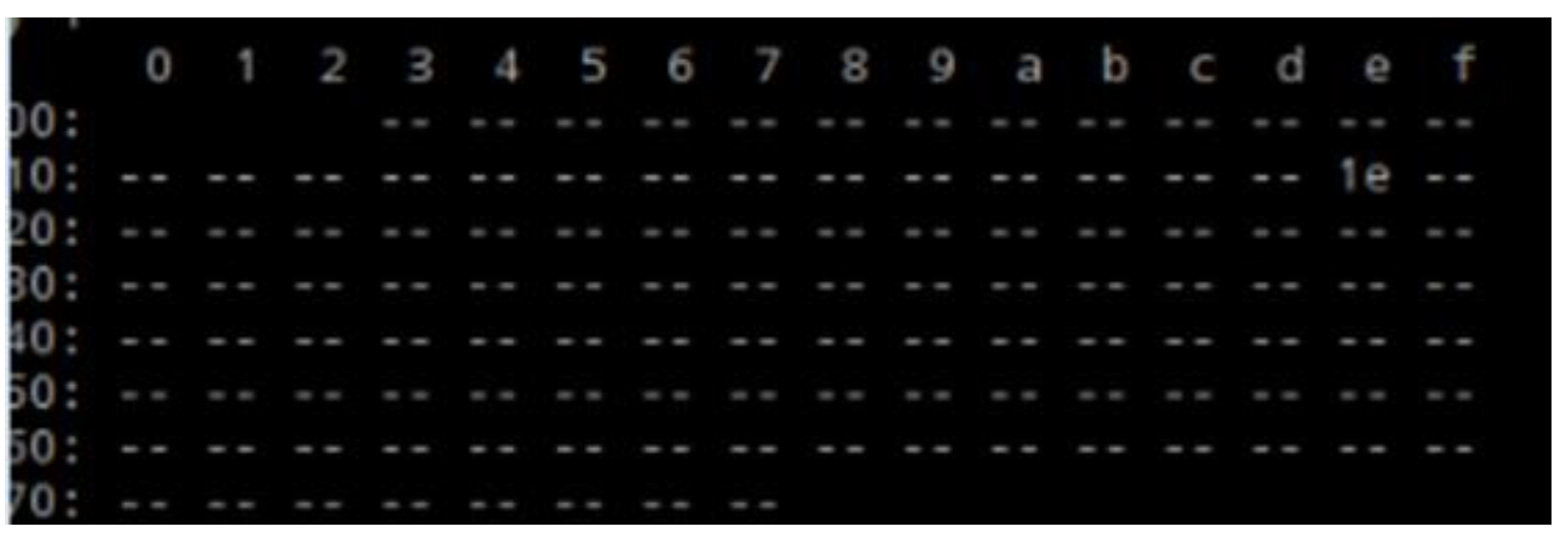}
  \caption{Compass Address}
  \label{fig:Compass}
\end{figure}

In the design of the hardware module compass SDA pin was connected to the Raspberry Pi board pin 3 and SCL was connected to board pin 6. To check the connection is correct command “i2cdetect -y 1” can be used as in this model of the raspberry pi used the port 1 of the i2c. Address of the compass as “1e” can experience on-grid blocks (Figure \ref{fig:Compass}).

Following each direction value in the path list, compass readings can be taken and if the reading is within the -/+2 range of the given value, a forward motion is taken by the robot. Else if the compass reading is not within the range continuous right and left motions are called until the module turn to the needed direction. Forward motion calling will move the robotic module from one cell to another. In the implementation 2x2 feet area was mapped as one cell in the array. So as the distance from one cell to another is fixed, the parameters for the forward motion was calculated as follows. The module has wheel encoders connected to the front two wheels and calculated the encoder tick count to move 2 feet distance. So, the forward motion call will execute until the given encoder ticks are reached.

\subsection{Obstacle avoidance}
Obstacle avoidance is provided by the ultrasonic sensor of the robot connected to the front of the module. Experiments were carried out to calculate the accuracy of the ultrasonic sensor values through a python program. The success rate of the ultrasonic sensors to detect objects was in the range of 100 cm. This will be given as a parameter in the avoidance process as the minimum distance to check for the capability of detecting an obstacle. 

If a dynamic obstacle is detected by the robot it will stop at that place and wait to check the presence of the obstacle. When the obstacle is away from the path of the robot again continues the usual procedure of key point’s coverage will happen. If the time to avoid the obstacle is high, then the navigation module first turns in to the right direction and check the presence of the obstacles in that direction. If the direction is free of obstacles a new path is generated without the key points that have been already covered. If the right move detects obstacles, then left move to the first direction is taken. If the left path is free of barriers, then again, a reconfiguration of the path will take place. Else if both right and left path are not free rotation to backward direction will take place and a new path will generate if the direction is free of obstacles. 

\section{Results}
The obtained results for the proposed solution can be elaborated as follows. A python GUI application was developed to track the path of the robotic platform and the following Figure \ref{fig:GUI} shows the robotic path where black dots represent the free spaces and blue dots represent the static obstacles in the environment.

\begin{figure}[ht]
  \centering
  \includegraphics[width=7.5cm, height=5.5cm]{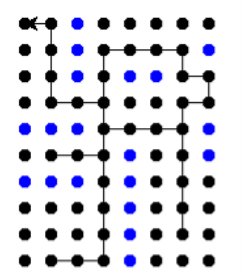}
  \caption{Keypoint coverage map}
  \label{fig:GUI}
\end{figure}

\begin{figure}[ht]
  \centering
  \includegraphics[width=10.5cm, height=5.5cm]{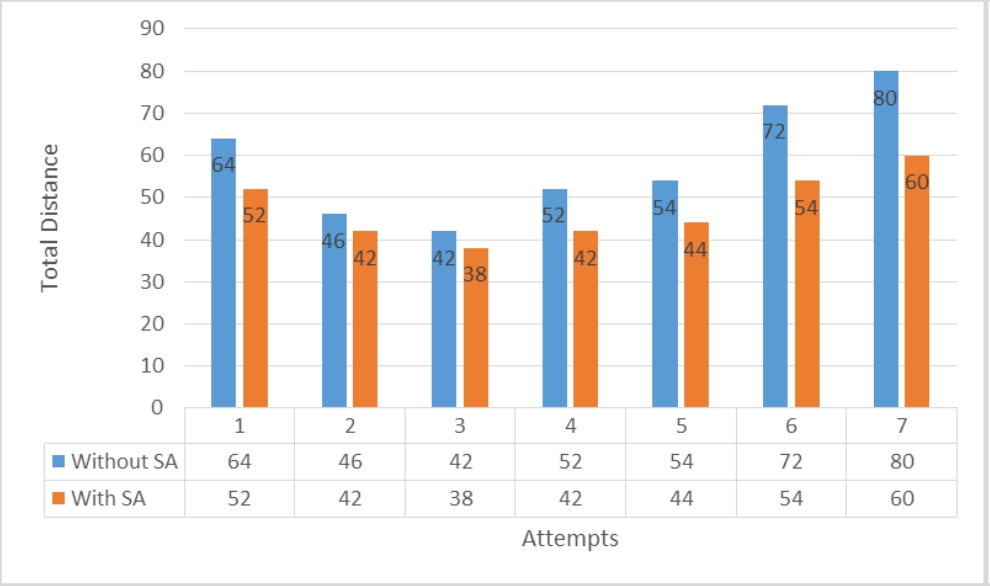}
  \caption{Total distance with and without SA}
  \label{fig:DistanceSA}
\end{figure}

In calculating the distance between each of two key points, Simulated Annealing (SA) process was used with the TSP algorithm to identify the global minima and then calculated the total distance. Reduction in the total distance can be obtained with the SA process and the shortest route can be identified as shown in Figure \ref{fig:DistanceSA}. 

Accuracy of the ultrasonic sensor was measured using a python code by keeping obstacles at 5, 10, 15, 20, 25, 50, 75,100 cm and 3, 4, 8 feet locations. The following Figure \ref{fig:Ultrasonic} will show data measured by the ultrasonic sensor. From the following graph, it can be noticed that the ultrasonic sensor values are accurate below the range of 100 cm.

\begin{figure}[ht]
  \centering
  \includegraphics[width=10.5cm, height=5.5cm]{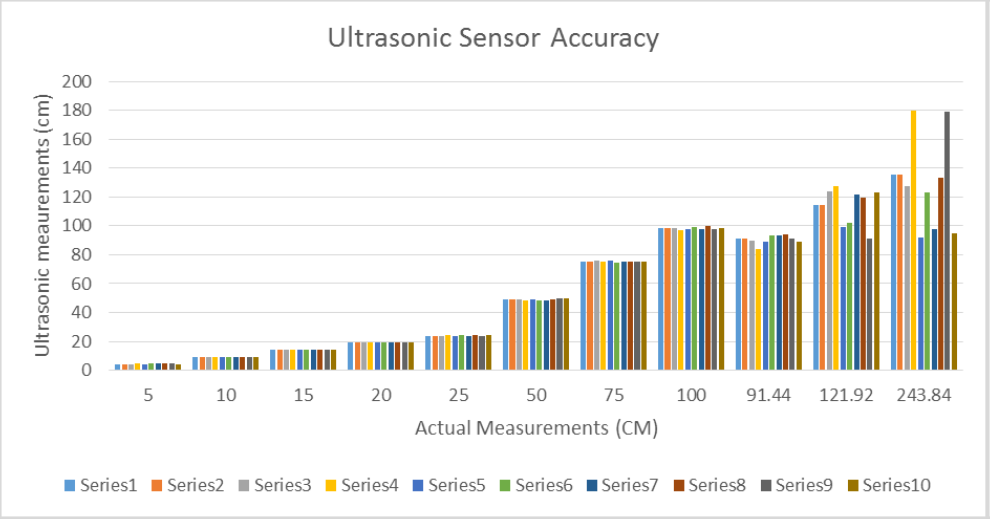}
  \caption{Ultrasonic sensor accuracy}
  \label{fig:Ultrasonic}
\end{figure}

Experiments were done to detect the ability of the robot to rotate until the given direction is reached and go for a forward motion in an even surface. The following Figure \ref{fig:ForwardMotion} shows the details of the experiment in which the forward Pulse Width Modulation (PWM) was set as 60 and right and left PWM were set as 30 each. The test was done to check the ability of the compass to select a bearing within the -/+2 range of the given bearing.

\begin{figure}[ht]
  \centering
  \includegraphics[width=10.5cm, height=5.5cm]{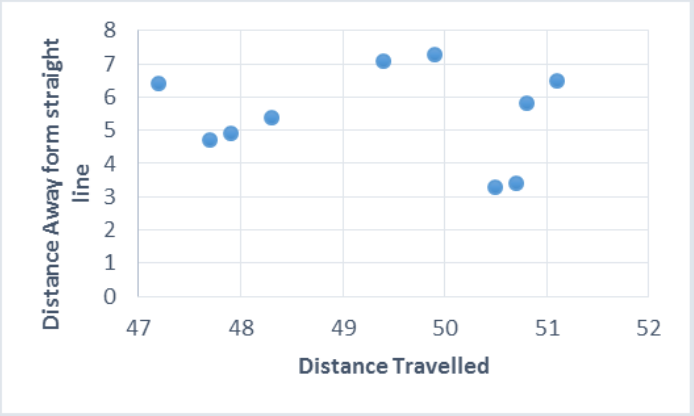}
  \caption{Forward motion ability}
  \label{fig:ForwardMotion}
\end{figure}

From the above chart, it is depicted that for a forward movement in the range 47cm to 51.5 cm at PWM 60, the distance travelled away from the straight line is within the range of 3cm to 8cm. 

\section{Discussion}
This paper suggests a solution to a common problem of path planning using a robotic platform operated by a raspberry pi module. This path planning approach focuses on the coverage of the specific key areas optimally with static and dynamic obstacle avoidance with the implementation details of the hardware platform. This approach required very few configurations to adapt to a new environment and can be easily implemented in any dynamic indoor situation.

\bibliographystyle{unsrt}

\end{document}